\PassOptionsToPackage{obeyspaces}{url}
\documentclass[sigconf,screen]{acmart}

\usepackage[utf8]{inputenc}
\usepackage[T1]{fontenc}
\usepackage{microtype}

\usepackage{bm}
\usepackage{soul}
\usepackage{graphicx}
\usepackage{amsmath,amsfonts}
\usepackage{booktabs}
\usepackage{multirow}
\usepackage{makecell}
\usepackage{textcomp}
\usepackage{array}
\usepackage{textcomp}
\usepackage[font=footnotesize]{subfig}
\usepackage{multirow}
\usepackage{float}
\usepackage{xspace}
\usepackage[export]{adjustbox}
\usepackage{nicefrac}
\usepackage{colortbl}
\usepackage{tablefootnote}
\usepackage[referable]{threeparttablex}
\usepackage{comment}
\usepackage{listings}
\usepackage{mathtools}
\usepackage{enumitem}

\usepackage{wrapfig}
\BeforeBeginEnvironment{wrapfigure}{\setlength{\intextsep}{2pt}}

\makeatletter
\@ifpackageloaded{icml\the\year}{
\usepackage{enumitem}
\setlist{nosep}}
{\usepackage[linesnumbered,ruled,vlined]{algorithm2e}
\SetKwRepeat{Do}{do}{while}
\DontPrintSemicolon

\SetCommentSty{commentsty}
}

\@ifclassloaded{IEEEtran}{
\usepackage[numbers,sort,compress]{natbib}

}{}

\@ifclassloaded{acmart}{
\def\authornotetext#1{
	\g@addto@macro\@authornotes{%
	\stepcounter{footnote}\footnotetext{#1}}%
}}{
\usepackage[usenames,svgnames]{xcolor}
\usepackage{hyperref}
\hypersetup{
    colorlinks = true,
    breaklinks = true,
    bookmarksdepth = section,
    citecolor = DarkGreen,
    linkcolor = DarkRed,
    urlcolor = DarkBlue,
}}

\@ifclassloaded{llncs}{
\usepackage[square,comma,numbers,sort&compress]{natbib}
\bibliographystyle{splncsnat}

}{
\usepackage{amsthm}

\theoremstyle{remark}

\theoremstyle{definition}

}
\makeatother

\usepackage{pifont}

\definecolor{sblue}{HTML}{20C1EE}
\definecolor{sred}{HTML}{E54F44}
\definecolor{spink}{HTML}{DE47A2}
\definecolor{sgreen}{HTML}{1AAF54}
\definecolor{spurple}{HTML}{7869BE}
\definecolor{syellow}{HTML}{EDB429}

\DeclarePairedDelimiterX{\infdivx}[2]{(}{)}{%
	#1\;\delimsize\|\;#2%
}

\DeclareMathAlphabet{\mathsfit}{\encodingdefault}{\sfdefault}{m}{sl}
\SetMathAlphabet{\mathsfit}{bold}{\encodingdefault}{\sfdefault}{bx}{n}

\newcommand{\header}[1]{{\vspace{+1mm}\flushleft \textbf{#1}}}

\copyrightyear{2022}
\acmYear{2022}
\setcopyright{acmcopyright}\acmConference[CIKM '22]{Proceedings of the 31st ACM International Conference on Information and Knowledge Management}{October 17--21, 2022}{Atlanta, GA, USA}
\acmBooktitle{Proceedings of the 31st ACM International Conference on Information and Knowledge Management (CIKM '22), October 17--21, 2022, Atlanta, GA, USA}
\acmPrice{15.00}
\acmDOI{10.1145/3511808.3557661}
\acmISBN{978-1-4503-9236-5/22/10}



\begin{document}

\title{On Positional and Structural Node Features \mbox{for Graph Neural Networks on Non-attributed Graphs}}

\author{Hejie Cui}
\affiliation{
\institution{Department of Computer Science, Emory University}
\city{Atlanta}
\state{GA}
\country{United States}
}
\email{hejie.cui@emory.edu}

\author{Zijie Lu}
\affiliation{
\institution{Department of Computer Science, University of Illinois at Urbana-Champaign}
\city{Champaign}
\state{IL}
\country{United States}
}
\email{zijielu2@illinois.edu}

\author{Pan Li}
\affiliation{
\institution{Department of Computer Science, Purdue University}
\city{West Lafayette}
\state{IN}
\country{United States}
}
\email{panli@purdue.edu}

\author{Carl Yang$^*$}
\affiliation{
\institution{Department of Computer Science, Emory University}
\city{Atlanta}
\state{GA}
\country{United States}
}
\email{j.carlyang@emory.edu}

\authornotetext{To whom correspondence should be addressed.}

\begin{abstract}
Graph neural networks (GNNs) have been widely used in various graph-related problems such as node classification and graph classification, where the superior performance is mainly established when natural node features are available. However, it is not well understood how GNNs work without natural node features, especially regarding the various ways to construct artificial ones. In this paper, we point out the two types of artificial node features, \textit{i.e.}, \textit{positional} and \textit{structural} node features, and provide insights on why each of them is more appropriate for certain tasks, \textit{i.e.}, \textit{positional node classification}, \textit{structural node classification}, and \textit{graph classification}. Extensive experimental results on 10 benchmark datasets validate our insights, thus leading to a practical guideline on the choices between different artificial node features for GNNs on non-attributed graphs. The code is available at \url{https://github.com/zjzijielu/gnn-positional-structural-node-features}.
\end{abstract}


\maketitle

\section{Introduction}
\label{sec:intro}
Graphs provide a concise yet rich representation of data across different domains such as social networks, citation networks, gene-protein interactions, molecular structures and so on. How to effectively mine valuable information underneath graph data has become an appealing problem for data mining community. Recently, various kinds of powerful Graph Neural Networks (GNNs) demonstrate their privilege on common graph tasks such as node classification \cite{kipf2016variational, DBLP:conf/bigdataconf/IzadiFSL20}, link prediction \cite{DBLP:conf/nips/ZhangC18,zhang2020revisiting,kan2021zero} and graph classification \cite{xu2018powerful, bacciu2018contextual, DBLP:conf/aaai/ZhangCNC18, liu2022size}. 
GNNs combine both node features and graph structures by aggregating node features through links into low-dimensional vector representations. 
Recently, considerable efforts have been put on studying the complicated contents of networks, such as node types and informativeness \cite{yang2020heterogeneous, DBLP:conf/nips/HuXQYC0T20}, pooling layers \cite{DBLP:conf/nips/MesquitaSK20, DBLP:conf/nips/LiC0T20}, design spaces \cite{DBLP:conf/nips/YouYL20, cui2022braingb}, heterogeneous graph \cite{liu2017semantic, zhu2022structure}, graph generation and transformation \cite{kan2022fbnetgen, wang2022deep, guo2022graph}, graph learning schema \cite{yang2022data}, task-specific GNNs \cite{cui2022interpretable} and so on, where the superior performances are mainly established when natural node features (i.e., attributes) are available when applying GNNs.

However, a great number of graphs in the wild do not contain node attributes \cite{chen2020learning, DBLP:journals/corr/abs-1911-08795}, which deteriorates the performance of GNNs \cite{errica_fair_2020, cai2018simple}. For example, in the molecules dataset QM9 \cite{DBLP:journals/jcisd/RuddigkeitDBR12, ramakrishnan2014quantum}, a graph represents a molecule, i.e., nodes are atoms and edges are chemical bonds. For typical tasks on this dataset such as predicting the properties of molecules, i.e., toxicity or biological activity, GNNs cannot be directly applied due to the lack of natural node features \cite{chen2020learning, DBLP:journals/fgcs/TaguchiLM21}. Another example is the social network such as REDDIT. In this dataset, each graph represents a discussion thread, where each node corresponds to one user, and two nodes are connected by an edge if one user responded to a comment of the other \cite{Morris+2020}. The missing of node features for each user in these social networks will introduce extra difficulties in the task of sub-reddits prediction.

\begin{figure}[t]
    \centering
    \includegraphics[width=\linewidth]{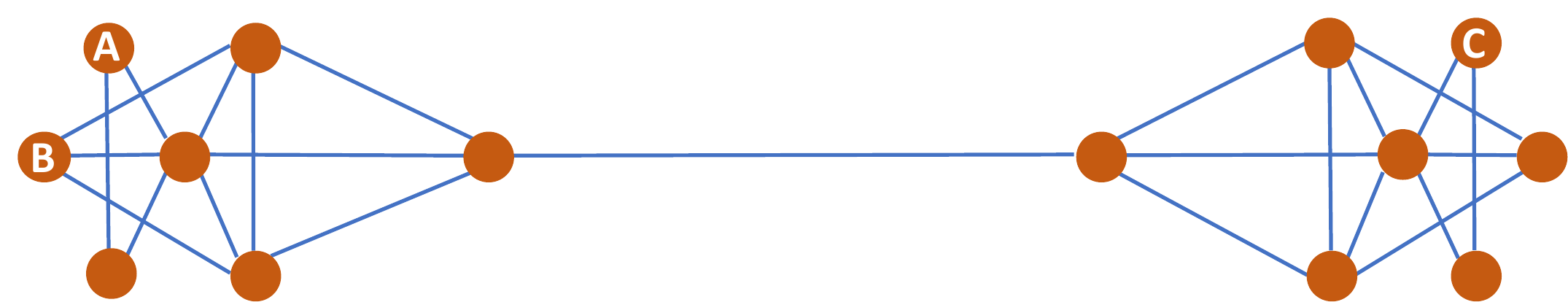}
    \caption{Illustration of Position vs. Structure: A and B are ``positionally close''-- having relatively close positions in the global network, whereas A and C are ``structurally close''-- having relatively similar local neighborhood structures.}
    \label{fig:toy-example}
\end{figure}

To apply GNNs on non-attributed graphs, several intuitive methods have been commonly practiced to initialize node features, such as degree-based~\cite{NIPS2017_5dd9db5e}, random~\cite{sato2021random,abboud2020surprising}, one-hot~\cite{chen2018supervised}, position-based~\cite{you2019position}, distance-based~\cite{li2020distance,you2021identity} and so on. However, to the best of our knowledge, there exists no generic understanding or guideline towards the initialization of artificial node features based on the needs of downstream tasks. In this paper, we categorize common artificial node features and study their utility towards different types of graph mining tasks. From a high level, these intuitive node feature initialization methods can be grouped into two categories, positional and structural ones \cite{chami2020machine} (Section \ref{sec:features}). Take Figure \ref{fig:toy-example} as an example. Positional features can help GNNs put node A and node B closer in the embedding space, whereas structural features facilitates putting node A and node C closer. 

Extensive experiments are performed on 10 datasets with 8 common artificial features. Based on the information needs of different tasks, we further categorize them into multiple divisions, namely, positional node classification, structural node classification and graph classification (Section \ref{sec:exp}). Observations on the results validate our understanding that positional node features are more suitable for positional node classification, while structural node features benefit more for structural node classification and graph classification tasks. With appropriately designed artificial node features, the performance of GNNs can even surpass that with real features in some cases, as indicated in Table \ref{tab:graph_result}. Besides, our proposed novel degree-based node feature initialization method, i.e., degree bucket range, achieves state-of-the-art performance on structural node classification (Section \ref{subsec:degree+}). We believe this empirical study on the selection of artificial node features can facilitate the understanding of feature initialization on non-attributed graphs and inspire new designs of artificial node features, thus shedding light on various GNN applications on graphs in the wild. 


\section{Two Types of Artificial Node Features}
\label{sec:features}
Several node feature initialization methods have been proposed for non-attributed graphs and commonly applied in various GNN models. We group these artificial node features into two main families: positional node features and structural node features. 

\subsection{Positional Node Features}
Positional node features help GNNs capture node distance information regarding their relative positions in the graph. For example, in Figure \ref{fig:toy-example}, nodes A and B are positionally close. A real case is the publication network, where two authors who cite each other and also cite / get cited by similar other authors should be close considering their graph positions, and recognized as sharing similar research interests.
Some intuitive positional node features include: 
\begin{itemize}[leftmargin=*]
    \item \textit{random}: A feature vector following random distribution is generated for each node, which is decided by the random seed in the data initialization. 
    The random feature of each node varies among training runs with different random seeds initialization.
    This feature itself does not reflect relative positions, but it records a high-dimensional identity for each node, which can indirectly help GNNs learn the relative node positions.
    \item \textit{one-hot}: A unique one-hot vector is initialized for each node \cite{errica_fair_2020, you2019position}. This feature is essentially equivalent to \textit{random}, when the first linear layer of the GNN are randomly initialized.
    \item \textit{eigen}: Eigen decomposition is performed on the normalized adjacency matrix and then the top \textit{k} eigen vectors are used to generate a \textit{k}-dimensional feature vector for each node \cite{DBLP:journals/corr/abs-2010-13993, chaudhuri2012spectral, DBLP:conf/nips/ZhangR18}, where the optimal value of \textit{k} is decided by grid search \cite{DBLP:journals/corr/abs-2105-03178}.
    \item \textit{deepwalk}: The initial feature of a node is generated based on the DeepWalk algorithm from \cite{perozzi2014deepwalk} with the walk length set as 40 by default. Deep walk features with walk length longer than 2 can help to capture higher-order positional information in the graph.
\end{itemize}
Correspondingly, positional node classifications target at grouping nodes with respect to their positions, which corresponds to coarse global information in the graph. For example, in Figure \ref{fig:toy-example}, nodes A and B should be classified into the same class in the task of positional node classification. Specifically, \textit{eigen} and \textit{deepwalk} methods which generate features by matrix decomposition \cite{DBLP:conf/wsdm/QiuDMLWT18}, are essentially dimension reduction, where the complex graph structures (i.e., adjacency matrices) information are embedded into a low dimensional representation. Therefore, \textit{eigen} and \textit{deepwalk} methods also incorporate structural information. However, as the features based on \textit{eigen} and \textit{deepwalk} reflect the position of nodes, with some abuse of terminology, we keep calling them positional features.


\subsection{Structural Node Features}
On the other hand, structural node features help GNNs capture structural information of nodes, such as degree information and neighborhood connection patterns. For example, in Figure \ref{fig:toy-example}, node A and C are similar regarding their neighborhood structures in the graph, though they are far away from each other in position. A real case is the molecular network, where two nodes with similar degrees and connection patterns should be put close considering their structures, and recognized as atoms with similar properties or functions.
Some intuitive node feature initialization methods focusing on the structural aspects include: 
\begin{itemize}[leftmargin=*]
    \item \textit{shared}: An initial feature vector is shared across all nodes \cite{errica_fair_2020}. The shared feature we used is simply a vector of all 1's. 
    \item \textit{degree}: An one-hot degree vector is initialized for each node, whose dimension is decided by the max degree of all nodes \cite{xu2018powerful, NIPS2017_5dd9db5e}.
    \item \textit{pagerank}: The original PageRank score \cite{DBLP:journals/cn/BrinP98} of a given node is calculated and then flattened into a vector in order to fully utilize the embedding dimensions of neural networks, where the dimension of the extended vector is selected by grid-search \cite{DBLP:journals/corr/abs-2105-03178}. It can be viewed as generalized higher-order node degree information. 
\end{itemize}
Structural node classifications target at classifying nodes according to their structural patterns. For example, nodes A and C in Figure \ref{fig:toy-example} should be put into the same class considering their similar ``structural roles''. Different from positional features that characterize the position of nodes in a graph, structural node features target at representing structural roles. Recent distance-based features~\cite{li2020distance,you2021identity} also help to learn node structural roles while GNNs that leverage distance-based features cannot make inference over multiple nodes in parallel, which increases the computational complexity. Interested readers may refer to the experiments in \cite{yin2020revisit} to check how distance-based features help with learning node structural roles.
 
\subsection{Byproduct: New SOTA for Structural Node Classification}
\label{subsec:degree+}
Motivated by our empirical studies on structural node features, we propose a novel node feature initialization method based on bucketing node degrees, which we name as \textit{degree+}. Specifically, we divide degree values into several buckets, then map the degree values distributed in each bucket range into one class, and finally construct a unique one-hot vector for each class. Our proposed \textit{degree+} feature can be regarded as an improved version of the original degree-based node feature, which better handles the sparse and skewed distribution of node degrees in the graph.

\section {Experimental Results} 
\label{sec:exp}
\subsection{Basic Settings}
To conduct a fair and unbiased evaluation on the effectiveness of node features, we adopt the popular GNN of GraphSAGE  \cite{DBLP:conf/nips/HamiltonYL17} with \textit{mean} and \textit{sum} aggregators for all the artificial node feature initialization methods across different types of graph mining tasks. Results with real features are also provided wherever natural node features are available. The train/test/validation split of each dataset follows the standard practice in the literature \cite{DBLP:conf/iclr/KipfW17, ribeiro2017struc2vec, errica_fair_2020}. Graph level experiments are conducted with artificial features of sizes ranging from 100 to 500 with step 100. In addition, we perform comprehensive grid search for the best hyper-parameter settings including the learning rate, number of epochs and neighborhood sample size. The final performance of each feature initialization method is averaged over five runs under the optimal hyper-parameter settings.

\subsection{Positional Node Classification}
\header{Definition and Datasets.} 
The tasks of positional node classification target at predicting the ``positional role'' of each node \cite{henderson2011s, DBLP:conf/iclr/Srinivasan020, liu2021relative}. We consider three datasets for positional node classification, including Cora \cite{sen2008collective}, Citeseer \cite{sen2008collective} and Pubmed \cite{namata:mlg12-wkshp}. These three citation networks consist of scientific publications as nodes, which can be classified into several content categories. Edges connecting those nodes denote the citation relationships between publications. Real features for each publication node are included in these three datasets, which are bag-of-word vectors indicating the word presence in the text content. In these three datasets, since the publications are connected by citation links, the research topic based node classification tasks should be mainly driven by the positions of nodes in the graph. Performance with the real node feature is also presented as a baseline for comparison. 
\header{Protocols and Performances.} 
We train and test the GraphSAGE model using the same data splits as in \cite{DBLP:conf/iclr/KipfW17}, namely 20 randomly-selected samples for each class during training with a validation set of 500 samples. Experiment results of different node features on three positional node classification datasets are presented in Table \ref{tab:positional_result}, where $\mathcal{P}$ and $\mathcal{S}$ indicate the \textit{Type} of artificial node features, corresponding to \textit{Positional} or \textit{Structural} respectively. \textit{Aggr.} denotes the aggregation method used in each GNN layer. Classification accuracy \textit{Acc.(\%)} is adopted here for evaluation.
\begin{table}[htbp]
    \centering
    \small
    \resizebox{0.4\textwidth}{!}{
    \begin{tabular}{cclccc}
        \toprule
        \multirow{2}{*}{Aggr.} & \multirow{2}{*}{Type} & \multirow{2}{*}{Feature} & \multicolumn{1}{c}{\textbf{Cora}} & \multicolumn{1}{c}{\textbf{Pubmed}} & \multicolumn{1}{c}{\textbf{Citeseer}}\\
        & & & \it Acc.(\%) & \it Acc.(\%)  & \it Acc.(\%)  \\
        \midrule
        \multirow{8}{*}{Mean}
        & \multirow{4}{*}{$\mathcal{P}$} 
        & random & 56.1$\pm$1.6 & 42.3$\pm$1.4 & 36.0$\pm$1.0 \\
        & & one-hot & 58.2$\pm$4.0 & 51.4$\pm$3.1 & 37.3$\pm$2.5 \\
        & & eigen & 73.2$\pm$2.3 & 70.0$\pm$4.8 & 42.9$\pm$2.3\\
        & & deepwalk & \textbf{75.3$\pm$1.0} & \textbf{74.0$\pm$2.6} & \textbf{46.8$\pm$0.9} \\
        \cmidrule{2-6}
        & \multirow{3}{*}{$\mathcal{S}$} 
        & shared & 17.9$\pm$0.0 & 38.6$\pm$0.0 & 20.2$\pm$0.0 \\
        & & degree  & 37.4$\pm$2.1 & 41.1$\pm$2.9 & 36.0$\pm$1.3 \\
        & & pagerank & 25.2$\pm$2.4 & 39.8$\pm$1.9 & 20.5$\pm$3.4\\
        \cmidrule{2-6}
        & & \cellcolor{gray!20} real feat. & \cellcolor{gray!20} 80.2$\pm$1.1 & \cellcolor{gray!20} 79.0$\pm$2.2 & \cellcolor{gray!20} 68.0$\pm$4.0\\
        \midrule
        \multirow{8}{*}{Sum} 
        & \multirow{4}{*}{$\mathcal{P}$} 
        & random & 45.2$\pm$3.9 & 41.7$\pm$2.7 & 32.8$\pm$2.7\\
        & & one-hot & 47.0$\pm$3.7 & 46.4$\pm$4.4 & 33.0$\pm$1.8\\
        & & eigen & 70.5$\pm$5.1 & 68.8$\pm$4.1 & 40.1$\pm$5.0\\
        & & deepwalk & 70.0$\pm$2.3 & 72.5$\pm$2.2 & 43.7$\pm$2.7\\
        \cmidrule{2-6}
        & \multirow{3}{*}{$\mathcal{S}$}
        & shared & 17.1$\pm$5.2 & 33.3$\pm$6.4 & 22.3$\pm$4.6\\
        & & degree & 50.7$\pm$3.7 & 42.6$\pm$1.8 & 32.0$\pm$3.5\\
        & & pagerank & 27.8$\pm$4.4 & 33.0$\pm$6.3 & 23.4$\pm$1.3\\
        \cmidrule{2-6}
        & & \cellcolor{gray!20} real feat. & \cellcolor{gray!20} 70.5$\pm$3.7 & \cellcolor{gray!20} 75.4$\pm$3.7 & \cellcolor{gray!20} 59.3$\pm$4.0\\
        \bottomrule
    \end{tabular}
    }
    \caption{Positional node classification results}
    \label{tab:positional_result}
\end{table}

\header{Observations.}
\begin{itemize}[leftmargin=*]
    \item \textit{Aggregation}: 
    For positional node classification, \textit{mean} aggregation shows better performance than \textit{sum} aggregation, since \textit{mean} aggregation can effectively filter out the influence of neighborhood size, which makes little contribution to and even impairs the performance on positional node classification. However, \textit{shared} feature plus \textit{mean} aggregation gives the same embedding for every node, so the results are constantly poor with no variance.
    \item \textit{Cross Feature Type Comparison}: 
    For positional node classification tasks, most positional node feature initialization methods achieve much better performance than structural node feature ones. The advantage of position node features over structural node features is especially remarkable with \textit{mean} aggregation. 
    \item \textit{Within Feature Type Comparison}: Among all positional node features: 1. \textit{random} and \textit{one-hot} achieve comparable results. This is because they are essentially the same: after passing through the first layer of neural network where the parameters are randomly initialized, one-hot initialization is equivalent to random initialization except for possible differences in dimensions (e.g., on Pubmed). 
    2. among all positional features, \textit{deepwalk} and \textit{eigen} demonstrate the best performance across all the datasets, which owes to the higher-order positional information they can capture. 
\end{itemize}

\subsection{Structural Node Classification}
\header{Definition and Datasets.} 
The tasks of structural node classification target at predicting the ``structural role'' of each node \cite{henderson2011s, henderson2012rolx, DBLP:journals/debu/HamiltonYL17}. Here we choose three datasets, namely American air-traffic network, Brazilian air-traffic network and European air-traffic network \cite{ribeiro2017struc2vec}. Given an airport node in the air-traffic network, the target is to predict passenger flow level of that node solely based on the structure of air-traffic network. These three datasets are chosen because the node labels of them indicate the structural roles (vary in four levels from hubs to switches), rather than the traditional community identifiers of nodes \cite{DBLP:conf/iclr/KipfW17, NIPS2017_5dd9db5e, sen2008collective}. 

\header{Protocols and Performances.} 
Following struc2vec \cite{ribeiro2017struc2vec}, we use 80\% of nodes for training. To highlight the performance of our novel \textit{degree+} method, we adopt logistic regression with L2 regularization to train the classifier using the representation learned by struc2vec \cite{ribeiro2017struc2vec}, which demonstrates SOTA results on these datasets. Experiment results are presented in Table \ref{tab:structural_result}. 

\begin{table}[t]
    \centering
    \small
    \resizebox{0.4\textwidth}{!}{
    \begin{tabular}{cclccc}
        \toprule
        \multirow{2}{*}{Aggr.} & \multirow{2}{*}{Type} & \multirow{2}{*}{Initial.} & \multicolumn{1}{c}{\textbf{USA-air}} & \multicolumn{1}{c}{\textbf{Brazil-air}} & \multicolumn{1}{c}{\textbf{Europe-air}}\\
        & & & \it Acc.(\%)  & \it Acc.(\%)   & \it Acc.(\%) \\
        \midrule
        \multirow{8}{*}{Mean} 
        & \multirow{4}{*}{$\mathcal{P}$} 
        & random & 59.3$\pm$1.8 & 45.7$\pm$5.9 & 44.9$\pm$5.8\\
        & & one-hot & 59.2$\pm$2.6 & 48.6$\pm$7.4 & 44.0$\pm$0.7\\
        & & eigen & 55.3$\pm$1.5 & 40.0$\pm$6.9 & 31.6$\pm$2.1\\
        & & deepwalk & 58.1$\pm$2.8 & 42.1$\pm$9.6 & 41.5$\pm$3.3 \\
        \cmidrule{2-6}
        & \multirow{4}{*}{$\mathcal{S}$} 
        & shared & 25.0$\pm$0.0 & 25.0$\pm$0.0 & 25.0$\pm$0.0 \\
        & & degree & 53.8$\pm$1.9 & 48.6$\pm$4.1 & 42.7$\pm$2.7 \\
        & & degree+ & 59.2$\pm$2.7 & 60.0$\pm$3.0 & 50.6$\pm$3.9\\
        & & pagerank & 39.7$\pm$2.9 & 47.9$\pm$7.4 & 25.9$\pm$0.0\\
        \midrule
        \multirow{8}{*}{Sum} 
        & \multirow{4}{*}{$\mathcal{P}$} 
        & random & 60.7$\pm$3.2 & 47.9$\pm$7.4 & 48.9$\pm$5.1\\
        & & one-hot & 59.2$\pm$3.3 & 50.7$\pm$8.5 & 48.9$\pm$5.4\\
        & & eigen & 67.8$\pm$2.5 & 57.8$\pm$5.3 & 49.4$\pm$4.5\\
        & & deepwalk & 68.8$\pm$3.0 & 65.0$\pm$6.4 & 54.1$\pm$2.8\\
        \cmidrule{2-6}
        & \multirow{4}{*}{$\mathcal{S}$} 
        & shared & 55.7$\pm$2.0 & 61.4$\pm$4.7 & 45.4$\pm$1.0\\
        & & degree & 63.6$\pm$3.0 & 70.0$\pm$4.1 & 58.0$\pm$3.6\\
        & & degree+ & \textbf{69.1$\pm$2.6} & \textbf{76.4$\pm$4.1} & \textbf{61.2$\pm$3.8} \\
        & & pagerank & 58.8$\pm$2.0 & 73.6$\pm$5.4 & 45.9$\pm$1.0\\
        \midrule
        \multirow{1}{*}{SOTA} 
        & & \cellcolor{gray!20} struc2vec  & \cellcolor{gray!20} 63.8$\pm$1.6 & \cellcolor{gray!20} 73.6$\pm$9.6  & \cellcolor{gray!20} 58.8$\pm$3.0\\
        \bottomrule
    \end{tabular}
    }
    \caption{Structural node classification results.}
    \label{tab:structural_result}
\end{table}

\header{Observations.}
\begin{itemize}[leftmargin=*]
    \item \textit{Aggregation}: 
    For structural node classification, \textit{sum} aggregation outperforms \textit{mean} aggregation because it can capture the number of neighbors, which is an important structural feature in graphs. 
    \item \textit{Cross Feature Type Comparison}: 
    For structural node classification tasks, in most cases structural node features demonstrate superiority compared with positional ones, and our proposed structural node feature \textit{degree+} manifests the most distinct advantage over other positional features, reaching the new state-of-the-art.
    \item \textit{Within Feature Type Comparison}: 
    1. among all four types of structural node features,  \textit{degree+} improves on \textit{degree} by using a degree bucket, where nodes with degree values in a range are projected together. This alleviates the node degree sparsity and skewness problem. 
    2. \textit{shared} can only capture the sizes of multi-hop neighborhoods, but loses track of neighborhood structures, thus performing rather poorly.
    3. In contrast, \textit{pagerank} can be viewed as a generalized higher-order node degree, and we conjecture that its performance deterioration arises from over-smoothing which in the worst cases renders it as similar to \textit{shared}.
\end{itemize}

\subsection{Graph Classification} 
\header{Definition and Datasets.} For graph classification, we consider two datasets with real node features, MUTAG \cite{debnath1991structure} and PROTEINS \cite{borgwardt2005protein} from chemical domain. We also include IMDB-BINARY and IMDB-MULTI \cite{yanardag2015deep} from the social domain without features.

\header{Protocols and Performances.} 
We take advantage of the GNN comparison framework proposed in \cite{errica_fair_2020}. On top of their experiment settings, we introduce the initialization methods, and use mean-  and sum-pooling when applying GraphSAGE for graph classification. 
\noindent Experiment results of different node initialization methods on graph classification datasets are presented in Table \ref{tab:graph_result}, where the \textit{real feat.} is only available for MUTAG and PROTEINS.
\begin{table}[t]
    \centering
    \footnotesize
    \resizebox{0.45\textwidth}{!}{
    \begin{tabular}{cclcccc}
        \toprule
        \multirow{2}{*}{Aggr.} & \multirow{2}{*}{Typ.} & \multirow{2}{*}{Initial.} & \multicolumn{1}{c}{\textbf{MUTAG}} & \multicolumn{1}{c}{\textbf{PROTEINS}} & \multicolumn{1}{c}{\textbf{IMDB-B}} & \multicolumn{1}{c}{\textbf{IMDB-M}}\\
        & & & \it Acc.(\%)  & \it Acc.(\%)  & \it Acc.(\%)  & \it Acc.(\%)   \\
        \midrule
        \multirow{8}{*}{Mean} 
        & \multirow{4}{*}{$\mathcal{P}$} 
        & random & 64.9$\pm$4.1 & 67.2$\pm$4.2 & 58.0$\pm$2.9 & 36.1$\pm$1.9\\
        & & one-hot & 65.8$\pm$7.0 & 67.8$\pm$2.6 & 56.9$\pm$3.4 & 36.8$\pm$3.2\\
        & & eigen & 63.8$\pm$2.1 & 60.4$\pm$1.0 & 50.2$\pm$1.3 & 33.4$\pm$0.7 \\
        & & deepwalk & 65.1$\pm$8.3 & 68.1$\pm$4.0 & 52.1$\pm$3.4 & 35.7$\pm$1.9\\
        \cmidrule{2-7}
        & \multirow{3}{*}{$\mathcal{S}$}
        & shared &  66.7$\pm$0.0 &  59.6$\pm$0.0 & 50.0$\pm$0.0 & 33.3$\pm$0.0\\
        & & degree & \textbf{84.4$\pm$7.7} & 69.5$\pm$2.6 & 69.7$\pm$5.1 & 45.1$\pm$ 2.6 \\
        & & pagerank & 66.5$\pm$1.9 & 68.0$\pm$5.5 & 54.4$\pm$4.0 & 35.5$\pm$1.7\\
        \cmidrule{2-7}
        & & \cellcolor{gray!20} real feat. & \cellcolor{gray!20} 71.4$\pm$4.4 & \cellcolor{gray!20} 74.0$\pm$4.2 & \cellcolor{gray!20} - & \cellcolor{gray!20} - \\
        \midrule
        \multirow{8}{*}{Sum} 
        & \multirow{4}{*}{$\mathcal{P}$} 
        & random & 66.9$\pm$7.1 & 67.5$\pm$4.1 & 54.0$\pm$3.6 & 36.2$\pm$2.1 \\
        & & one-hot & 65.1$\pm$3.8 & 66.8$\pm$3.8 & 52.8$\pm$2.7 & 33.4$\pm$2.6 \\
        & & eigen & 65.4$\pm$7.7 & 69.0$\pm$4.1 & 69.3$\pm$4.6 & 42.4$\pm$3.4 \\
        & & deepwalk & 64.2$\pm$8.6 & 66.2$\pm$4.2 & 51.9$\pm$2.8 & 35.3$\pm$3.0\\
        \cmidrule{2-7}
        & \multirow{3}{*}{$\mathcal{S}$} 
        & shared & 79.9$\pm$6.7 & 69.1$\pm$4.5 & 67.9$\pm$2.8 & 43.3$\pm$4.6  \\
        & & degree & 84.0$\pm$8.4 & 69.3$\pm$3.3 & 68.9$\pm$2.5 & 44.9$\pm$4.1\\
        & & pagerank & 77.3$\pm$7.6 & \textbf{69.9$\pm$3.1} & \textbf{70.3$\pm$2.9} & \textbf{48.2$\pm$3.2} \\
        \cmidrule{2-7}
        & & \cellcolor{gray!20} real feat. & \cellcolor{gray!20} 83.0$\pm$6.3 & \cellcolor{gray!20} 73.8$\pm$2.6 & \cellcolor{gray!20} - & \cellcolor{gray!20} - \\
        \bottomrule
    \end{tabular}
    }
    \caption{Graph classification results.}
    \label{tab:graph_result}
\end{table}

\header{Observations.}
\begin{itemize}[leftmargin=*]
    \item \textit{Aggregation}:
    Similar to structural classification tasks, \textit{sum} aggregation outperforms \textit{mean} aggregation on graph classification tasks, since the number of neighbors contributes as an important type of structural information for graph classification tasks. 
    \item \textit{Cross Feature Type Comparison}: 
    For graph classification, though the best performance is not consistently achieved on a particular feature across four datasets, it always falls in the category of structural ones. This is because we do not care about positional information such as the specific position of each node in graph classification. Instead, similar to structural node classification, the overall structural information of the graph matters. 
    \item \textit{Within Feature Type Comparison}: 
    1. Among the structural node features, \textit{pagerank} demonstrates better performance in most of the cases. 2. Impressively, the performances of GNN on \textit{degree} on MUATG and \textit{pagerank} on PROTEIN with the \textit{sum} aggregator even surpass those with real features. This further demonstrates the importance of choosing the appropriate artificial node features, sometimes even when natural node features are available.
\end{itemize}
\section{Conclusion}
\label{sec:con}
Graphs in the real world do not always have natural node features available, due to the lack of task-specific node attributes, privacy concerns and/or difficulties in data collection. 
In this paper, we study the usage of artificial node features when applying GNNs on non-attributed graphs. 
We categorize commonly used artificial node features into two groups, positional node features and structural node features, based on what kind of information they can help GNNs capture. Extensive empirical experiments are conducted across three graph mining tasks, positional node classification, structural node classification and graph classification. 
The results validate our insights that positional node features are more suitable for positional node classification, while structural node features benefit more for structural node classification and graph classification tasks.
We hope our empirical study can provide a generic and practical guideline for choosing the appropriate artificial node features and exploring more useful ones based on the needs of downstream tasks.

\balance
\bibliographystyle{ACM-Reference-Format}
\bibliography{cikm22}


\begin{thebibliography}{59}


\ifx \showCODEN    \undefined \def \showCODEN     #1{\unskip}     \fi
\ifx \showDOI      \undefined \def \showDOI       #1{#1}\fi
\ifx \showISBNx    \undefined \def \showISBNx     #1{\unskip}     \fi
\ifx \showISBNxiii \undefined \def \showISBNxiii  #1{\unskip}     \fi
\ifx \showISSN     \undefined \def \showISSN      #1{\unskip}     \fi
\ifx \showLCCN     \undefined \def \showLCCN      #1{\unskip}     \fi
\ifx \shownote     \undefined \def \shownote      #1{#1}          \fi
\ifx \showarticletitle \undefined \def \showarticletitle #1{#1}   \fi
\ifx \showURL      \undefined \def \showURL       {\relax}        \fi
\providecommand\bibfield[2]{#2}
\providecommand\bibinfo[2]{#2}
\providecommand\natexlab[1]{#1}
\providecommand\showeprint[2][]{arXiv:#2}

\bibitem[\protect\citeauthoryear{Abboud, Ceylan, Grohe, and Lukasiewicz}{Abboud
  et~al\mbox{.}}{2020}]%
        {abboud2020surprising}
\bibfield{author}{\bibinfo{person}{Ralph Abboud},
  \bibinfo{person}{{\.I}smail~{\.I}lkan Ceylan}, \bibinfo{person}{Martin
  Grohe}, {and} \bibinfo{person}{Thomas Lukasiewicz}.}
  \bibinfo{year}{2020}\natexlab{}.
\newblock \showarticletitle{The Surprising Power of Graph Neural Networks with
  Random Node Initialization}.
\newblock \bibinfo{journal}{\emph{arXiv preprint arXiv:2010.01179}}
  (\bibinfo{year}{2020}).
\newblock


\bibitem[\protect\citeauthoryear{Bacciu, Errica, and Micheli}{Bacciu
  et~al\mbox{.}}{2018}]%
        {bacciu2018contextual}
\bibfield{author}{\bibinfo{person}{Davide Bacciu}, \bibinfo{person}{Federico
  Errica}, {and} \bibinfo{person}{Alessio Micheli}.}
  \bibinfo{year}{2018}\natexlab{}.
\newblock \showarticletitle{Contextual graph markov model: A deep and
  generative approach to graph processing}. In
  \bibinfo{booktitle}{\emph{ICML}}.
\newblock


\bibitem[\protect\citeauthoryear{Borgwardt, Ong, Sch{\"o}nauer, Vishwanathan,
  Smola, and Kriegel}{Borgwardt et~al\mbox{.}}{2005}]%
        {borgwardt2005protein}
\bibfield{author}{\bibinfo{person}{Karsten~M Borgwardt},
  \bibinfo{person}{Cheng~Soon Ong}, \bibinfo{person}{Stefan Sch{\"o}nauer},
  \bibinfo{person}{SVN Vishwanathan}, \bibinfo{person}{Alex~J Smola}, {and}
  \bibinfo{person}{Hans-Peter Kriegel}.} \bibinfo{year}{2005}\natexlab{}.
\newblock \showarticletitle{Protein function prediction via graph kernels}.
\newblock \bibinfo{journal}{\emph{Bioinformatics}}  \bibinfo{volume}{21}
  (\bibinfo{year}{2005}), \bibinfo{pages}{i47--i56}.
\newblock


\bibitem[\protect\citeauthoryear{Brin and Page}{Brin and Page}{1998}]%
        {DBLP:journals/cn/BrinP98}
\bibfield{author}{\bibinfo{person}{Sergey Brin} {and} \bibinfo{person}{Lawrence
  Page}.} \bibinfo{year}{1998}\natexlab{}.
\newblock \showarticletitle{The Anatomy of a Large-Scale Hypertextual Web
  Search Engine}.
\newblock \bibinfo{journal}{\emph{Comput. Networks}}  \bibinfo{volume}{30}
  (\bibinfo{year}{1998}), \bibinfo{pages}{107--117}.
\newblock


\bibitem[\protect\citeauthoryear{Cai and Wang}{Cai and Wang}{2019}]%
        {cai2018simple}
\bibfield{author}{\bibinfo{person}{Chen Cai} {and} \bibinfo{person}{Yusu
  Wang}.} \bibinfo{year}{2019}\natexlab{}.
\newblock \showarticletitle{A simple yet effective baseline for non-attribute
  graph classification}.
\newblock \bibinfo{journal}{\emph{ICLR Workshop on Representation Learning on
  Graphs and Manifolds}} (\bibinfo{year}{2019}).
\newblock


\bibitem[\protect\citeauthoryear{Chami, Abu-El-Haija, Perozzi, R{\'e}, and
  Murphy}{Chami et~al\mbox{.}}{2020}]%
        {chami2020machine}
\bibfield{author}{\bibinfo{person}{Ines Chami}, \bibinfo{person}{Sami
  Abu-El-Haija}, \bibinfo{person}{Bryan Perozzi}, \bibinfo{person}{Christopher
  R{\'e}}, {and} \bibinfo{person}{Kevin Murphy}.}
  \bibinfo{year}{2020}\natexlab{}.
\newblock \showarticletitle{Machine learning on graphs: A model and
  comprehensive taxonomy}.
\newblock \bibinfo{journal}{\emph{arXiv preprint arXiv:2005.03675}}
  (\bibinfo{year}{2020}).
\newblock


\bibitem[\protect\citeauthoryear{Chaudhuri, Chung, and Tsiatas}{Chaudhuri
  et~al\mbox{.}}{2012}]%
        {chaudhuri2012spectral}
\bibfield{author}{\bibinfo{person}{Kamalika Chaudhuri}, \bibinfo{person}{Fan
  Chung}, {and} \bibinfo{person}{Alexander Tsiatas}.}
  \bibinfo{year}{2012}\natexlab{}.
\newblock \showarticletitle{Spectral clustering of graphs with general degrees
  in the extended planted partition model}. In
  \bibinfo{booktitle}{\emph{Conference on Learning Theory}}.
\newblock


\bibitem[\protect\citeauthoryear{Chen, Chen, Yao, Zheng, Zhang, and Tsang}{Chen
  et~al\mbox{.}}{2020}]%
        {chen2020learning}
\bibfield{author}{\bibinfo{person}{Xu Chen}, \bibinfo{person}{Siheng Chen},
  \bibinfo{person}{Jiangchao Yao}, \bibinfo{person}{Huangjie Zheng},
  \bibinfo{person}{Ya Zhang}, {and} \bibinfo{person}{Ivor~W Tsang}.}
  \bibinfo{year}{2020}\natexlab{}.
\newblock \showarticletitle{Learning on Attribute-Missing Graphs}.
\newblock \bibinfo{journal}{\emph{IEEE transactions on pattern analysis and
  machine intelligence}} (\bibinfo{year}{2020}).
\newblock


\bibitem[\protect\citeauthoryear{Chen, Li, and Bruna}{Chen
  et~al\mbox{.}}{2018}]%
        {chen2018supervised}
\bibfield{author}{\bibinfo{person}{Zhengdao Chen}, \bibinfo{person}{Lisha Li},
  {and} \bibinfo{person}{Joan Bruna}.} \bibinfo{year}{2018}\natexlab{}.
\newblock \showarticletitle{Supervised Community Detection with Line Graph
  Neural Networks}. In \bibinfo{booktitle}{\emph{ICLR}}.
\newblock


\bibitem[\protect\citeauthoryear{Cui, Dai, Zhu, Kan, Gu, Lukemire, Zhan, He,
  Guo, and Yang}{Cui et~al\mbox{.}}{2022a}]%
        {cui2022braingb}
\bibfield{author}{\bibinfo{person}{Hejie Cui}, \bibinfo{person}{Wei Dai},
  \bibinfo{person}{Yanqiao Zhu}, \bibinfo{person}{Xuan Kan},
  \bibinfo{person}{Antonio Aodong~Chen Gu}, \bibinfo{person}{Joshua Lukemire},
  \bibinfo{person}{Liang Zhan}, \bibinfo{person}{Lifang He},
  \bibinfo{person}{Ying Guo}, {and} \bibinfo{person}{Carl Yang}.}
  \bibinfo{year}{2022}\natexlab{a}.
\newblock \showarticletitle{BrainGB: A Benchmark for Brain Network Analysis
  with Graph Neural Networks}.
\newblock \bibinfo{journal}{\emph{arXiv preprint arXiv:2204.07054}}
  (\bibinfo{year}{2022}).
\newblock


\bibitem[\protect\citeauthoryear{Cui, Dai, Zhu, Li, He, and Yang}{Cui
  et~al\mbox{.}}{2022b}]%
        {cui2022interpretable}
\bibfield{author}{\bibinfo{person}{Hejie Cui}, \bibinfo{person}{Wei Dai},
  \bibinfo{person}{Yanqiao Zhu}, \bibinfo{person}{Xiaoxiao Li},
  \bibinfo{person}{Lifang He}, {and} \bibinfo{person}{Carl Yang}.}
  \bibinfo{year}{2022}\natexlab{b}.
\newblock \showarticletitle{Interpretable Graph Neural Networks for
  Connectome-Based Brain Disorder Analysis}. In
  \bibinfo{booktitle}{\emph{MICCAI}}.
\newblock


\bibitem[\protect\citeauthoryear{Debnath, Lopez~de Compadre, Debnath,
  Shusterman, and Hansch}{Debnath et~al\mbox{.}}{1991}]%
        {debnath1991structure}
\bibfield{author}{\bibinfo{person}{Asim~Kumar Debnath}, \bibinfo{person}{Rosa~L
  Lopez~de Compadre}, \bibinfo{person}{Gargi Debnath}, \bibinfo{person}{Alan~J
  Shusterman}, {and} \bibinfo{person}{Corwin Hansch}.}
  \bibinfo{year}{1991}\natexlab{}.
\newblock \showarticletitle{Structure-activity relationship of mutagenic
  aromatic and heteroaromatic nitro compounds. correlation with molecular
  orbital energies and hydrophobicity}.
\newblock \bibinfo{journal}{\emph{Journal of medicinal chemistry}}
  \bibinfo{volume}{34} (\bibinfo{year}{1991}), \bibinfo{pages}{786--797}.
\newblock


\bibitem[\protect\citeauthoryear{Duong, Hoang, Dang, Nguyen, and Aberer}{Duong
  et~al\mbox{.}}{2019}]%
        {DBLP:journals/corr/abs-1911-08795}
\bibfield{author}{\bibinfo{person}{Chi~Thang Duong}, \bibinfo{person}{Thanh~Dat
  Hoang}, \bibinfo{person}{Ha~The~Hien Dang}, \bibinfo{person}{Quoc Viet~Hung
  Nguyen}, {and} \bibinfo{person}{Karl Aberer}.}
  \bibinfo{year}{2019}\natexlab{}.
\newblock \showarticletitle{On Node Features for Graph Neural Networks}.
\newblock \bibinfo{journal}{\emph{CoRR}} (\bibinfo{year}{2019}).
\newblock


\bibitem[\protect\citeauthoryear{Errica, Podda, Bacciu, and Micheli}{Errica
  et~al\mbox{.}}{2020}]%
        {errica_fair_2020}
\bibfield{author}{\bibinfo{person}{Federico Errica}, \bibinfo{person}{Marco
  Podda}, \bibinfo{person}{Davide Bacciu}, {and} \bibinfo{person}{Alessio
  Micheli}.} \bibinfo{year}{2020}\natexlab{}.
\newblock \showarticletitle{A fair comparison of graph neural networks for
  graph classification}. In \bibinfo{booktitle}{\emph{ICLR}}.
\newblock


\bibitem[\protect\citeauthoryear{Guo, Wang, and Zhao}{Guo
  et~al\mbox{.}}{2022}]%
        {guo2022graph}
\bibfield{author}{\bibinfo{person}{Xiaojie Guo}, \bibinfo{person}{Shiyu Wang},
  {and} \bibinfo{person}{Liang Zhao}.} \bibinfo{year}{2022}\natexlab{}.
\newblock \showarticletitle{Graph Neural Networks: Graph Transformation}.
\newblock In \bibinfo{booktitle}{\emph{Graph Neural Networks: Foundations,
  Frontiers, and Applications}}. \bibinfo{pages}{251--275}.
\newblock


\bibitem[\protect\citeauthoryear{Hamilton, Ying, and Leskovec}{Hamilton
  et~al\mbox{.}}{2017a}]%
        {NIPS2017_5dd9db5e}
\bibfield{author}{\bibinfo{person}{Will Hamilton}, \bibinfo{person}{Zhitao
  Ying}, {and} \bibinfo{person}{Jure Leskovec}.}
  \bibinfo{year}{2017}\natexlab{a}.
\newblock \showarticletitle{Inductive Representation Learning on Large Graphs}.
  In \bibinfo{booktitle}{\emph{NeurIPS}}.
\newblock


\bibitem[\protect\citeauthoryear{Hamilton, Ying, and Leskovec}{Hamilton
  et~al\mbox{.}}{2017b}]%
        {DBLP:journals/debu/HamiltonYL17}
\bibfield{author}{\bibinfo{person}{William~L. Hamilton}, \bibinfo{person}{Rex
  Ying}, {and} \bibinfo{person}{Jure Leskovec}.}
  \bibinfo{year}{2017}\natexlab{b}.
\newblock \showarticletitle{Representation Learning on Graphs: Methods and
  Applications}.
\newblock \bibinfo{journal}{\emph{IEEE Data Engineering Bulletin}}
  \bibinfo{volume}{40} (\bibinfo{year}{2017}), \bibinfo{pages}{52--74}.
\newblock


\bibitem[\protect\citeauthoryear{Hamilton, Ying, and Leskovec}{Hamilton
  et~al\mbox{.}}{2017c}]%
        {DBLP:conf/nips/HamiltonYL17}
\bibfield{author}{\bibinfo{person}{William~L. Hamilton},
  \bibinfo{person}{Zhitao Ying}, {and} \bibinfo{person}{Jure Leskovec}.}
  \bibinfo{year}{2017}\natexlab{c}.
\newblock \showarticletitle{Inductive Representation Learning on Large Graphs}.
  In \bibinfo{booktitle}{\emph{NeurIPS}}.
\newblock


\bibitem[\protect\citeauthoryear{Henderson, Gallagher, Eliassi-Rad, Tong, Basu,
  Akoglu, Koutra, Faloutsos, and Li}{Henderson et~al\mbox{.}}{2012}]%
        {henderson2012rolx}
\bibfield{author}{\bibinfo{person}{Keith Henderson}, \bibinfo{person}{Brian
  Gallagher}, \bibinfo{person}{Tina Eliassi-Rad}, \bibinfo{person}{Hanghang
  Tong}, \bibinfo{person}{Sugato Basu}, \bibinfo{person}{Leman Akoglu},
  \bibinfo{person}{Danai Koutra}, \bibinfo{person}{Christos Faloutsos}, {and}
  \bibinfo{person}{Lei Li}.} \bibinfo{year}{2012}\natexlab{}.
\newblock \showarticletitle{Rolx: structural role extraction \& mining in large
  graphs}. In \bibinfo{booktitle}{\emph{SIGKDD}}.
\newblock


\bibitem[\protect\citeauthoryear{Henderson, Gallagher, Li, Akoglu, Eliassi-Rad,
  Tong, and Faloutsos}{Henderson et~al\mbox{.}}{2011}]%
        {henderson2011s}
\bibfield{author}{\bibinfo{person}{Keith Henderson}, \bibinfo{person}{Brian
  Gallagher}, \bibinfo{person}{Lei Li}, \bibinfo{person}{Leman Akoglu},
  \bibinfo{person}{Tina Eliassi-Rad}, \bibinfo{person}{Hanghang Tong}, {and}
  \bibinfo{person}{Christos Faloutsos}.} \bibinfo{year}{2011}\natexlab{}.
\newblock \showarticletitle{It's who you know: graph mining using recursive
  structural features}. In \bibinfo{booktitle}{\emph{SIGKDD}}.
\newblock


\bibitem[\protect\citeauthoryear{Hu, Xiong, Qu, Yuan, C{\^{o}}t{\'{e}}, Liu,
  and Tang}{Hu et~al\mbox{.}}{2020}]%
        {DBLP:conf/nips/HuXQYC0T20}
\bibfield{author}{\bibinfo{person}{Shengding Hu}, \bibinfo{person}{Zheng
  Xiong}, \bibinfo{person}{Meng Qu}, \bibinfo{person}{Xingdi Yuan},
  \bibinfo{person}{Marc{-}Alexandre C{\^{o}}t{\'{e}}}, \bibinfo{person}{Zhiyuan
  Liu}, {and} \bibinfo{person}{Jian Tang}.} \bibinfo{year}{2020}\natexlab{}.
\newblock \showarticletitle{Graph Policy Network for Transferable Active
  Learning on Graphs}. In \bibinfo{booktitle}{\emph{NeurIPS}}.
\newblock


\bibitem[\protect\citeauthoryear{Huang, He, Singh, Lim, and Benson}{Huang
  et~al\mbox{.}}{2020}]%
        {DBLP:journals/corr/abs-2010-13993}
\bibfield{author}{\bibinfo{person}{Qian Huang}, \bibinfo{person}{Horace He},
  \bibinfo{person}{Abhay Singh}, \bibinfo{person}{Ser{-}Nam Lim}, {and}
  \bibinfo{person}{Austin~R. Benson}.} \bibinfo{year}{2020}\natexlab{}.
\newblock \showarticletitle{Combining Label Propagation and Simple Models
  Out-performs Graph Neural Networks}.
\newblock \bibinfo{journal}{\emph{CoRR}}  \bibinfo{volume}{abs/2010.13993}
  (\bibinfo{year}{2020}).
\newblock


\bibitem[\protect\citeauthoryear{Izadi, Fang, Stevenson, and Lin}{Izadi
  et~al\mbox{.}}{2020}]%
        {DBLP:conf/bigdataconf/IzadiFSL20}
\bibfield{author}{\bibinfo{person}{Mohammad~Rasool Izadi},
  \bibinfo{person}{Yihao Fang}, \bibinfo{person}{Robert Stevenson}, {and}
  \bibinfo{person}{Lizhen Lin}.} \bibinfo{year}{2020}\natexlab{}.
\newblock \showarticletitle{Optimization of Graph Neural Networks with Natural
  Gradient Descent}. In \bibinfo{booktitle}{\emph{{IEEE} BigData}}.
  \bibinfo{pages}{171--179}.
\newblock


\bibitem[\protect\citeauthoryear{Kan, Cui, Lukemire, Guo, and Yang}{Kan
  et~al\mbox{.}}{2022}]%
        {kan2022fbnetgen}
\bibfield{author}{\bibinfo{person}{Xuan Kan}, \bibinfo{person}{Hejie Cui},
  \bibinfo{person}{Joshua Lukemire}, \bibinfo{person}{Ying Guo}, {and}
  \bibinfo{person}{Carl Yang}.} \bibinfo{year}{2022}\natexlab{}.
\newblock \showarticletitle{Fbnetgen: Task-aware gnn-based fmri analysis via
  functional brain network generation}. In \bibinfo{booktitle}{\emph{MIDL}}.
\newblock


\bibitem[\protect\citeauthoryear{Kan, Cui, and Yang}{Kan et~al\mbox{.}}{2021}]%
        {kan2021zero}
\bibfield{author}{\bibinfo{person}{Xuan Kan}, \bibinfo{person}{Hejie Cui},
  {and} \bibinfo{person}{Carl Yang}.} \bibinfo{year}{2021}\natexlab{}.
\newblock \showarticletitle{Zero-shot scene graph relation prediction through
  commonsense knowledge integration}. In \bibinfo{booktitle}{\emph{ECML-PKDD}}.
\newblock


\bibitem[\protect\citeauthoryear{Kipf and Welling}{Kipf and Welling}{2016}]%
        {kipf2016variational}
\bibfield{author}{\bibinfo{person}{Thomas~N Kipf} {and} \bibinfo{person}{Max
  Welling}.} \bibinfo{year}{2016}\natexlab{}.
\newblock \showarticletitle{Variational graph auto-encoders}. In
  \bibinfo{booktitle}{\emph{NIPS Workshop on Bayesian Deep Learning}}.
\newblock


\bibitem[\protect\citeauthoryear{Kipf and Welling}{Kipf and Welling}{2017}]%
        {DBLP:conf/iclr/KipfW17}
\bibfield{author}{\bibinfo{person}{Thomas~N. Kipf} {and} \bibinfo{person}{Max
  Welling}.} \bibinfo{year}{2017}\natexlab{}.
\newblock \showarticletitle{Semi-Supervised Classification with Graph
  Convolutional Networks}. In \bibinfo{booktitle}{\emph{ICLR}}.
\newblock


\bibitem[\protect\citeauthoryear{Li, Chen, Zhang, and Tsang}{Li
  et~al\mbox{.}}{2020a}]%
        {DBLP:conf/nips/LiC0T20}
\bibfield{author}{\bibinfo{person}{Maosen Li}, \bibinfo{person}{Siheng Chen},
  \bibinfo{person}{Ya Zhang}, {and} \bibinfo{person}{Ivor~W. Tsang}.}
  \bibinfo{year}{2020}\natexlab{a}.
\newblock \showarticletitle{Graph Cross Networks with Vertex Infomax Pooling}.
  In \bibinfo{booktitle}{\emph{NeurIPS}}.
\newblock


\bibitem[\protect\citeauthoryear{Li, Wang, Wang, and Leskovec}{Li
  et~al\mbox{.}}{2020b}]%
        {li2020distance}
\bibfield{author}{\bibinfo{person}{Pan Li}, \bibinfo{person}{Yanbang Wang},
  \bibinfo{person}{Hongwei Wang}, {and} \bibinfo{person}{Jure Leskovec}.}
  \bibinfo{year}{2020}\natexlab{b}.
\newblock \showarticletitle{Distance Encoding: Design Provably More Powerful
  Neural Networks for Graph Representation Learning}.
\newblock \bibinfo{journal}{\emph{NeurIPS}} (\bibinfo{year}{2020}).
\newblock


\bibitem[\protect\citeauthoryear{Liu, Fang, Liu, and Hoi}{Liu
  et~al\mbox{.}}{2021}]%
        {liu2021relative}
\bibfield{author}{\bibinfo{person}{Zemin Liu}, \bibinfo{person}{Yuan Fang},
  \bibinfo{person}{Chenghao Liu}, {and} \bibinfo{person}{Steven~CH Hoi}.}
  \bibinfo{year}{2021}\natexlab{}.
\newblock \showarticletitle{Relative and absolute location embedding for
  few-shot node classification on graph}. In \bibinfo{booktitle}{\emph{AAAI}}.
\newblock


\bibitem[\protect\citeauthoryear{Liu, Mao, Liu, Fang, and Sun}{Liu
  et~al\mbox{.}}{2022}]%
        {liu2022size}
\bibfield{author}{\bibinfo{person}{Zemin Liu}, \bibinfo{person}{Qiheng Mao},
  \bibinfo{person}{Chenghao Liu}, \bibinfo{person}{Yuan Fang}, {and}
  \bibinfo{person}{Jianling Sun}.} \bibinfo{year}{2022}\natexlab{}.
\newblock \showarticletitle{On Size-Oriented Long-Tailed Graph Classification
  of Graph Neural Networks}. In \bibinfo{booktitle}{\emph{WWW}}.
  \bibinfo{pages}{1506--1516}.
\newblock


\bibitem[\protect\citeauthoryear{Liu, Zheng, Zhao, Zhu, Chang, Wu, and
  Ying}{Liu et~al\mbox{.}}{2017}]%
        {liu2017semantic}
\bibfield{author}{\bibinfo{person}{Zemin Liu}, \bibinfo{person}{Vincent~W
  Zheng}, \bibinfo{person}{Zhou Zhao}, \bibinfo{person}{Fanwei Zhu},
  \bibinfo{person}{Kevin Chen-Chuan Chang}, \bibinfo{person}{Minghui Wu}, {and}
  \bibinfo{person}{Jing Ying}.} \bibinfo{year}{2017}\natexlab{}.
\newblock \showarticletitle{Semantic proximity search on heterogeneous graph by
  proximity embedding}. In \bibinfo{booktitle}{\emph{AAAI}}.
\newblock


\bibitem[\protect\citeauthoryear{Luo, Li, Peng, Yang, Sun, Yu, and He}{Luo
  et~al\mbox{.}}{2021}]%
        {DBLP:journals/corr/abs-2105-03178}
\bibfield{author}{\bibinfo{person}{Gongxu Luo}, \bibinfo{person}{Jianxin Li},
  \bibinfo{person}{Hao Peng}, \bibinfo{person}{Carl Yang},
  \bibinfo{person}{Lichao Sun}, \bibinfo{person}{Philip~S. Yu}, {and}
  \bibinfo{person}{Lifang He}.} \bibinfo{year}{2021}\natexlab{}.
\newblock \showarticletitle{Graph Entropy Guided Node Embedding Dimension
  Selection for Graph Neural Networks}.
\newblock \bibinfo{journal}{\emph{CoRR}}  \bibinfo{volume}{abs/2105.03178}
  (\bibinfo{year}{2021}).
\newblock


\bibitem[\protect\citeauthoryear{Mesquita, Jr., and Kaski}{Mesquita
  et~al\mbox{.}}{2020}]%
        {DBLP:conf/nips/MesquitaSK20}
\bibfield{author}{\bibinfo{person}{Diego P.~P. Mesquita},
  \bibinfo{person}{Amauri H.~Souza Jr.}, {and} \bibinfo{person}{Samuel Kaski}.}
  \bibinfo{year}{2020}\natexlab{}.
\newblock \showarticletitle{Rethinking pooling in graph neural networks}. In
  \bibinfo{booktitle}{\emph{NeurIPS}}.
\newblock


\bibitem[\protect\citeauthoryear{Morris, Kriege, Bause, Kersting, Mutzel, and
  Neumann}{Morris et~al\mbox{.}}{2020}]%
        {Morris+2020}
\bibfield{author}{\bibinfo{person}{Christopher Morris},
  \bibinfo{person}{Nils~M. Kriege}, \bibinfo{person}{Franka Bause},
  \bibinfo{person}{Kristian Kersting}, \bibinfo{person}{Petra Mutzel}, {and}
  \bibinfo{person}{Marion Neumann}.} \bibinfo{year}{2020}\natexlab{}.
\newblock \showarticletitle{TUDataset: A collection of benchmark datasets for
  learning with graphs}. In \bibinfo{booktitle}{\emph{ICML Workshop on Graph
  Representation Learning and Beyond (GRL+ 2020)}}.
\newblock


\bibitem[\protect\citeauthoryear{Namata, London, Getoor, and Huang}{Namata
  et~al\mbox{.}}{2012}]%
        {namata:mlg12-wkshp}
\bibfield{author}{\bibinfo{person}{Galileo~Mark Namata}, \bibinfo{person}{Ben
  London}, \bibinfo{person}{Lise Getoor}, {and} \bibinfo{person}{Bert Huang}.}
  \bibinfo{year}{2012}\natexlab{}.
\newblock \showarticletitle{Query-driven Active Surveying for Collective
  Classification}. In \bibinfo{booktitle}{\emph{Workshop on Mining and Learning
  with Graphs}}.
\newblock


\bibitem[\protect\citeauthoryear{Perozzi, Al-Rfou, and Skiena}{Perozzi
  et~al\mbox{.}}{2014}]%
        {perozzi2014deepwalk}
\bibfield{author}{\bibinfo{person}{Bryan Perozzi}, \bibinfo{person}{Rami
  Al-Rfou}, {and} \bibinfo{person}{Steven Skiena}.}
  \bibinfo{year}{2014}\natexlab{}.
\newblock \showarticletitle{Deepwalk: Online learning of social
  representations}. In \bibinfo{booktitle}{\emph{SIGKDD}}.
\newblock


\bibitem[\protect\citeauthoryear{Qiu, Dong, Ma, Li, Wang, and Tang}{Qiu
  et~al\mbox{.}}{2018}]%
        {DBLP:conf/wsdm/QiuDMLWT18}
\bibfield{author}{\bibinfo{person}{Jiezhong Qiu}, \bibinfo{person}{Yuxiao
  Dong}, \bibinfo{person}{Hao Ma}, \bibinfo{person}{Jian Li},
  \bibinfo{person}{Kuansan Wang}, {and} \bibinfo{person}{Jie Tang}.}
  \bibinfo{year}{2018}\natexlab{}.
\newblock \showarticletitle{Network Embedding as Matrix Factorization: Unifying
  DeepWalk, LINE, PTE, and node2vec}. In \bibinfo{booktitle}{\emph{{WSDM}}}.
\newblock


\bibitem[\protect\citeauthoryear{Ramakrishnan, Dral, Rupp, and von
  Lilienfeld}{Ramakrishnan et~al\mbox{.}}{2014}]%
        {ramakrishnan2014quantum}
\bibfield{author}{\bibinfo{person}{Raghunathan Ramakrishnan},
  \bibinfo{person}{Pavlo~O Dral}, \bibinfo{person}{Matthias Rupp}, {and}
  \bibinfo{person}{O~Anatole von Lilienfeld}.} \bibinfo{year}{2014}\natexlab{}.
\newblock \showarticletitle{Quantum chemistry structures and properties of 134
  kilo molecules}.
\newblock \bibinfo{journal}{\emph{Scientific Data}}  \bibinfo{volume}{1}
  (\bibinfo{year}{2014}).
\newblock


\bibitem[\protect\citeauthoryear{Ribeiro, Saverese, and Figueiredo}{Ribeiro
  et~al\mbox{.}}{2017}]%
        {ribeiro2017struc2vec}
\bibfield{author}{\bibinfo{person}{Leonardo~FR Ribeiro},
  \bibinfo{person}{Pedro~HP Saverese}, {and} \bibinfo{person}{Daniel~R
  Figueiredo}.} \bibinfo{year}{2017}\natexlab{}.
\newblock \showarticletitle{struc2vec: Learning node representations from
  structural identity}. In \bibinfo{booktitle}{\emph{SIGKDD}}.
\newblock


\bibitem[\protect\citeauthoryear{Ruddigkeit, van Deursen, Blum, and
  Reymond}{Ruddigkeit et~al\mbox{.}}{2012}]%
        {DBLP:journals/jcisd/RuddigkeitDBR12}
\bibfield{author}{\bibinfo{person}{Lars Ruddigkeit}, \bibinfo{person}{Ruud van
  Deursen}, \bibinfo{person}{Lorenz~C. Blum}, {and}
  \bibinfo{person}{Jean{-}Louis Reymond}.} \bibinfo{year}{2012}\natexlab{}.
\newblock \showarticletitle{Enumeration of 166 Billion Organic Small Molecules
  in the Chemical Universe Database {GDB-17}}.
\newblock \bibinfo{journal}{\emph{J. Chem. Inf. Model.}}  \bibinfo{volume}{52}
  (\bibinfo{year}{2012}), \bibinfo{pages}{2864--2875}.
\newblock


\bibitem[\protect\citeauthoryear{Sato, Yamada, and Kashima}{Sato
  et~al\mbox{.}}{2021}]%
        {sato2021random}
\bibfield{author}{\bibinfo{person}{Ryoma Sato}, \bibinfo{person}{Makoto
  Yamada}, {and} \bibinfo{person}{Hisashi Kashima}.}
  \bibinfo{year}{2021}\natexlab{}.
\newblock \showarticletitle{Random features strengthen graph neural networks}.
  In \bibinfo{booktitle}{\emph{SIAM International Conference on Data Mining
  (SDM)}}.
\newblock


\bibitem[\protect\citeauthoryear{Sen, Namata, Bilgic, Getoor, Galligher, and
  Eliassi-Rad}{Sen et~al\mbox{.}}{2008}]%
        {sen2008collective}
\bibfield{author}{\bibinfo{person}{Prithviraj Sen}, \bibinfo{person}{Galileo
  Namata}, \bibinfo{person}{Mustafa Bilgic}, \bibinfo{person}{Lise Getoor},
  \bibinfo{person}{Brian Galligher}, {and} \bibinfo{person}{Tina Eliassi-Rad}.}
  \bibinfo{year}{2008}\natexlab{}.
\newblock \showarticletitle{Collective classification in network data}.
\newblock \bibinfo{journal}{\emph{AI magazine}}  \bibinfo{volume}{29}
  (\bibinfo{year}{2008}), \bibinfo{pages}{93--93}.
\newblock


\bibitem[\protect\citeauthoryear{Srinivasan and Ribeiro}{Srinivasan and
  Ribeiro}{2020}]%
        {DBLP:conf/iclr/Srinivasan020}
\bibfield{author}{\bibinfo{person}{Balasubramaniam Srinivasan} {and}
  \bibinfo{person}{Bruno Ribeiro}.} \bibinfo{year}{2020}\natexlab{}.
\newblock \showarticletitle{On the Equivalence between Positional Node
  Embeddings and Structural Graph Representations}. In
  \bibinfo{booktitle}{\emph{{ICLR}}}.
\newblock


\bibitem[\protect\citeauthoryear{Taguchi, Liu, and Murata}{Taguchi
  et~al\mbox{.}}{2021}]%
        {DBLP:journals/fgcs/TaguchiLM21}
\bibfield{author}{\bibinfo{person}{Hibiki Taguchi}, \bibinfo{person}{Xin Liu},
  {and} \bibinfo{person}{Tsuyoshi Murata}.} \bibinfo{year}{2021}\natexlab{}.
\newblock \showarticletitle{Graph convolutional networks for graphs containing
  missing features}.
\newblock \bibinfo{journal}{\emph{Future Gener. Comput. Syst.}}
  \bibinfo{volume}{117} (\bibinfo{year}{2021}), \bibinfo{pages}{155--168}.
\newblock


\bibitem[\protect\citeauthoryear{Wang, Guo, and Zhao}{Wang
  et~al\mbox{.}}{2022}]%
        {wang2022deep}
\bibfield{author}{\bibinfo{person}{Shiyu Wang}, \bibinfo{person}{Xiaojie Guo},
  {and} \bibinfo{person}{Liang Zhao}.} \bibinfo{year}{2022}\natexlab{}.
\newblock \showarticletitle{Deep Generative Model for Periodic Graphs}.
\newblock \bibinfo{journal}{\emph{arXiv preprint arXiv:2201.11932}}
  (\bibinfo{year}{2022}).
\newblock


\bibitem[\protect\citeauthoryear{Xu, Hu, Leskovec, and Jegelka}{Xu
  et~al\mbox{.}}{2019}]%
        {xu2018powerful}
\bibfield{author}{\bibinfo{person}{Keyulu Xu}, \bibinfo{person}{Weihua Hu},
  \bibinfo{person}{Jure Leskovec}, {and} \bibinfo{person}{Stefanie Jegelka}.}
  \bibinfo{year}{2019}\natexlab{}.
\newblock \showarticletitle{How Powerful are Graph Neural Networks?}. In
  \bibinfo{booktitle}{\emph{ICLR}}.
\newblock


\bibitem[\protect\citeauthoryear{Yanardag and Vishwanathan}{Yanardag and
  Vishwanathan}{2015}]%
        {yanardag2015deep}
\bibfield{author}{\bibinfo{person}{Pinar Yanardag} {and} \bibinfo{person}{SVN
  Vishwanathan}.} \bibinfo{year}{2015}\natexlab{}.
\newblock \showarticletitle{Deep graph kernels}. In
  \bibinfo{booktitle}{\emph{SIGKDD}}.
\newblock


\bibitem[\protect\citeauthoryear{Yang, Xiao, Zhang, Sun, and Han}{Yang
  et~al\mbox{.}}{2020}]%
        {yang2020heterogeneous}
\bibfield{author}{\bibinfo{person}{Carl Yang}, \bibinfo{person}{Yuxin Xiao},
  \bibinfo{person}{Yu Zhang}, \bibinfo{person}{Yizhou Sun}, {and}
  \bibinfo{person}{Jiawei Han}.} \bibinfo{year}{2020}\natexlab{}.
\newblock \showarticletitle{Heterogeneous Network Representation Learning: A
  Unified Framework with Survey and Benchmark}.
\newblock \bibinfo{journal}{\emph{TKDE}} (\bibinfo{year}{2020}).
\newblock


\bibitem[\protect\citeauthoryear{Yang, Zhu, Cui, Kan, He, Guo, and Yang}{Yang
  et~al\mbox{.}}{2022}]%
        {yang2022data}
\bibfield{author}{\bibinfo{person}{Yi Yang}, \bibinfo{person}{Yanqiao Zhu},
  \bibinfo{person}{Hejie Cui}, \bibinfo{person}{Xuan Kan},
  \bibinfo{person}{Lifang He}, \bibinfo{person}{Ying Guo}, {and}
  \bibinfo{person}{Carl Yang}.} \bibinfo{year}{2022}\natexlab{}.
\newblock \showarticletitle{Data-Efficient Brain Connectome Analysis via
  Multi-Task Meta-Learning}. In \bibinfo{booktitle}{\emph{KDD}}.
\newblock


\bibitem[\protect\citeauthoryear{Yin, Wang, and Li}{Yin et~al\mbox{.}}{2020}]%
        {yin2020revisit}
\bibfield{author}{\bibinfo{person}{Haoteng Yin}, \bibinfo{person}{Yanbang
  Wang}, {and} \bibinfo{person}{Pan Li}.} \bibinfo{year}{2020}\natexlab{}.
\newblock \showarticletitle{Revisit graph neural networks and distance encoding
  in a practical view}.
\newblock \bibinfo{journal}{\emph{arXiv preprint arXiv:2011.12228}}
  (\bibinfo{year}{2020}).
\newblock


\bibitem[\protect\citeauthoryear{You, Gomes-Selman, Ying, and Leskovec}{You
  et~al\mbox{.}}{2021}]%
        {you2021identity}
\bibfield{author}{\bibinfo{person}{Jiaxuan You}, \bibinfo{person}{Jonathan
  Gomes-Selman}, \bibinfo{person}{Rex Ying}, {and} \bibinfo{person}{Jure
  Leskovec}.} \bibinfo{year}{2021}\natexlab{}.
\newblock \showarticletitle{Identity-Aware Graph Neural Networks}. In
  \bibinfo{booktitle}{\emph{AAAI}}.
\newblock


\bibitem[\protect\citeauthoryear{You, Ying, and Leskovec}{You
  et~al\mbox{.}}{2019}]%
        {you2019position}
\bibfield{author}{\bibinfo{person}{Jiaxuan You}, \bibinfo{person}{Rex Ying},
  {and} \bibinfo{person}{Jure Leskovec}.} \bibinfo{year}{2019}\natexlab{}.
\newblock \showarticletitle{Position-aware graph neural networks}. In
  \bibinfo{booktitle}{\emph{ICML}}.
\newblock


\bibitem[\protect\citeauthoryear{You, Ying, and Leskovec}{You
  et~al\mbox{.}}{2020}]%
        {DBLP:conf/nips/YouYL20}
\bibfield{author}{\bibinfo{person}{Jiaxuan You}, \bibinfo{person}{Zhitao Ying},
  {and} \bibinfo{person}{Jure Leskovec}.} \bibinfo{year}{2020}\natexlab{}.
\newblock \showarticletitle{Design Space for Graph Neural Networks}. In
  \bibinfo{booktitle}{\emph{NeurIPS}}.
\newblock


\bibitem[\protect\citeauthoryear{Zhang and Chen}{Zhang and Chen}{2018}]%
        {DBLP:conf/nips/ZhangC18}
\bibfield{author}{\bibinfo{person}{Muhan Zhang} {and} \bibinfo{person}{Yixin
  Chen}.} \bibinfo{year}{2018}\natexlab{}.
\newblock \showarticletitle{Link Prediction Based on Graph Neural Networks}. In
  \bibinfo{booktitle}{\emph{NeurIPS}}.
\newblock


\bibitem[\protect\citeauthoryear{Zhang, Cui, Neumann, and Chen}{Zhang
  et~al\mbox{.}}{2018}]%
        {DBLP:conf/aaai/ZhangCNC18}
\bibfield{author}{\bibinfo{person}{Muhan Zhang}, \bibinfo{person}{Zhicheng
  Cui}, \bibinfo{person}{Marion Neumann}, {and} \bibinfo{person}{Yixin Chen}.}
  \bibinfo{year}{2018}\natexlab{}.
\newblock \showarticletitle{An End-to-End Deep Learning Architecture for Graph
  Classification}. In \bibinfo{booktitle}{\emph{{AAAI}}}.
\newblock


\bibitem[\protect\citeauthoryear{Zhang, Li, Xia, Wang, and Jin}{Zhang
  et~al\mbox{.}}{2020}]%
        {zhang2020revisiting}
\bibfield{author}{\bibinfo{person}{Muhan Zhang}, \bibinfo{person}{Pan Li},
  \bibinfo{person}{Yinglong Xia}, \bibinfo{person}{Kai Wang}, {and}
  \bibinfo{person}{Long Jin}.} \bibinfo{year}{2020}\natexlab{}.
\newblock \showarticletitle{Revisiting Graph Neural Networks for Link
  Prediction}.
\newblock \bibinfo{journal}{\emph{arXiv preprint arXiv:2010.16103}}
  (\bibinfo{year}{2020}).
\newblock


\bibitem[\protect\citeauthoryear{Zhang and Rohe}{Zhang and Rohe}{2018}]%
        {DBLP:conf/nips/ZhangR18}
\bibfield{author}{\bibinfo{person}{Yilin Zhang} {and} \bibinfo{person}{Karl
  Rohe}.} \bibinfo{year}{2018}\natexlab{}.
\newblock \showarticletitle{Understanding Regularized Spectral Clustering via
  Graph Conductance}. In \bibinfo{booktitle}{\emph{NeurIPS}}.
\newblock


\bibitem[\protect\citeauthoryear{Zhu, Xu, Cui, Yang, Liu, and Wu}{Zhu
  et~al\mbox{.}}{2022}]%
        {zhu2022structure}
\bibfield{author}{\bibinfo{person}{Yanqiao Zhu}, \bibinfo{person}{Yichen Xu},
  \bibinfo{person}{Hejie Cui}, \bibinfo{person}{Carl Yang},
  \bibinfo{person}{Qiang Liu}, {and} \bibinfo{person}{Shu Wu}.}
  \bibinfo{year}{2022}\natexlab{}.
\newblock \showarticletitle{Structure-enhanced heterogeneous graph contrastive
  learning}. In \bibinfo{booktitle}{\emph{SDM}}.
\newblock


\end{thebibliography}

\end{document}